# Spatial-Temporal Map Vehicle Trajectory Detection Using Dynamic Mode Decomposition and Res-UNet+ Neural Networks.

Tianya T. Zhang, *Student Member*, *IEEE*, Peter J. Jin, Ph.D. *Member*, *IEEE*

***Abstract*— This paper presents a machine-learning-enhanced longitudinal scanline method to extract vehicle trajectories from high-angle traffic cameras. The Dynamic Mode Decomposition (DMD) method is applied to extract vehicle strands by decomposing the Spatial-Temporal Map (STMap) into the sparse foreground and low-rank background. A deep neural network named Res-UNet+ was designed for the semantic segmentation task by adapting two prevalent deep learning architectures. The Res-UNet+ neural networks significantly improve the performance of the STMap-based vehicle detection, and the DMD model provides many interesting insights for understanding the evolution of underlying spatial-temporal structures preserved by STMap. The model outputs were compared with the previous image processing model and mainstream semantic segmentation deep neural networks. After a thorough evaluation, the model is proved to be accurate and robust against many challenging factors. Last but not least, this paper fundamentally addressed many quality issues found in NGSIM trajectory data. The cleaned high-quality trajectory data are published to support future theoretical and modeling research on traffic flow and microscopic vehicle control. This method is a reliable solution for video-based trajectory extraction and has wide applicability.**

## I. Introduction

Video sensor has been used widely to extract vehicle trajectory data to support academic research, traffic operations, management, and design. One of the most impactful video-based trajectory datasets is the next generation simulation (NGSIM) trajectory dataset [1], which has significantly boosted the traffic flow and modeling research by revealing microscopic traffic characteristics. As highlighted by [2], although video-based trajectory data have greatly improved the types of models and the calibration/training of the models, there is still a substantial demand for high-quality, high-resolution trajectory data. Collecting useful trajectory data from traffic cameras with satisfactory accuracy is a very challenging task. The traditional trajectory extraction paradigm, which contains multi-stage algorithms, is error-prone to the varying weather, illumination, video quality, and other factors. Vehicle trajectory dataset often requires significant efforts for post-processing, such as cleaning and validation.

To solve the data quality issues in the NGSIM dataset and meet the needs of traffic flow and modeling research, this paper presented a machine learning enhanced scanline method to detect trajectory from Spatial-Temporal Map (STMap). The Dynamic Mode Decomposition (DMD) is applied to analyze the STMap by decomposing it with different underlying structures. The DMD results were used to preprocess and prepare training data for a new deep neural network. Two existing convolutional neural network architectures were leveraged to build the Res-UNet+ model for the STMap segmentation task. This method has largely improved the scanline algorithm for vehicle detection and simplified high-fidelity trajectory data acquisition.

Furthermore, the paper also provides an efficient way to validate extracted vehicle trajectories by showing vehicle movements on the static STMap. The previous trajectory validation process relies on the direct approach by visualizing the detection and tracking results on raw video, or the indirect method by calibrating the microscopic model against both the raw and reconstructed trajectories and compare which of the two calibrated models is better. However, the spatial-temporal map allows identifying the error for the individual vehicle with minimal effort directly.

This paper is organized as follows. The related work section describes the state of the art of video analytics for vehicle trajectory extraction. The methodology section presents the proposed methods, including dynamic mode decomposition and Res-UNet+ convolutional network for STMap trajectory detection. Next, the experimental design section describes the dataset and model configurations, and the result analysis section analyzes the performance of the proposed models compared to baseline models. The final section concludes the paper and addresses future works.

## II. Related Work

### A. High-Resolution Vehicle Trajectory Datasets

Three significant high-resolution vehicle trajectory datasets and their processing methods are reviewed in this section.

**NGSIM:** NGSIM trajectory dataset, which is a great asset to the transportation research community, are generated with traffic videos taken from high-rise buildings. NGSIM applied an appearance-based vehicle detection algorithm to extract line segments from images and match them to 3D vehicle models. The detected vehicles were tracked according to their appearance in the camera image. The NGSIM dataset has been used to calibrate and evaluate traffic flow models as ground-truth data, demonstrate driving behavior or traffic phenomena, and conduct traffic-state estimation and prediction [3][4].

Submitted January 22nd, 2021. Revised May 31st, 2021. Accepted Oct 1st, 2021.

T. T. Zhang is a PhD candidate at the Civil and Environmental Engineering Department in Rutgers University, Piscataway, NJ 08854 USA (e-mail: tz140@soe.rutgers.edu).

P. J. Jin, is an Associate Professor at Civil and Environmental Engineering Department in Rutgers University, Piscataway, NJ 08854 USA (e-mail: peter.j.jin@rutgers.edu)

However, a growing amount of literature has uncovered the underlying systematic errors in the NGSIM dataset. Some research [5][6][7][8][9] examined the performance issue and proposed denoising methods based on statistic distributions, filtering & smoothing, traffic-informed constraints, and information theory [10].

**HighD & inD Dataset:** Krajewski et al. [11] published a HighD dataset that consists of 110,500 vehicles collected by Drones on German highways. The same group published an inD trajectory dataset of intersection road users containing pedestrians, bikers, vans, and others. The object detection algorithm used to generate HighD dataset is a U-Net semantic segmentation algorithm.

**pNEUMA**: In contrast to the highway vehicle trajectory dataset, Barmpounakis and Geroliminis [12] presented a complete urban dataset collected from a swamp of drones named pNEUMA (New Era of Urban traffic Monitoring with Aerial footage). Their project was conducted with a commercial traffic platform [13]. The dataset covers a congested area with more than 100 km-lanes and around 100 intersections under the multi-modal traffic environment using ten drones. The raw pNEUMA dataset does not contain lane information, which requires users to apply an additional lane identification method.

### *B. Computer Vision Algorithms for Traffic Video Analysis*

Traffic detection is a part of the object detection problem in computer vision. Great progress has been made in recent years with the rising of deep learning. Object detection involves not only recognizing the objects in targeted classes but precisely localizing each object. Table 1 is a summary of computer vision techniques related to traffic detection which can be classified into five major categories, including shape-based methods [14][15][16], background/foreground modeling [17][18][19], deep learning model [20][21][22][23], feature-based model [24][25], and scanline methods [26][27].

Semantic segmentation is another computer vision task related to traffic video analytics, which predicts class labels at the pixel level for each image. The challenges lie in the requirements of pixel-level accuracy and multi-scale contextual information of the class labels [28]. Semantic segmentation has been used in many applications: self-driving cars, virtual and augmented reality, biomedical image segmentation, etc. Many segmentation models are built upon prevailing neural networks, such as AlexNet [29], VGG-16 [30], GoogLeNet [31], and ResNet [32]. U-Net was first proposed as a semantic segmentation method in 2015 to process biomedical images [33]. The original vanilla U-Net has many variations with similar U-shape architecture, leading to a series of models as the U-Net model family.

TABLE I
VEHICLE DETECTION METHODOLOGIES AND ALGORITHMS

| Categories | Typical Algorithms |
|---|---|
| Shape Based Method [13 ~15] | Deformable Part-Based Model<br>Eigen Vehicle Model |
| Feature Based Methods [24,25] | SIFT/Surf<br>Harr-like<br>Optical Flow |
| Background Modeling [17 ~ 19] | Frame Difference<br>Approximate Median<br>Mixture of Gaussian |
| Deep Learning Model [20~23] | Fast/Faster R-CNN<br>Mask R-CNN<br>YOLO v1 ~ v5. |
| Scanline Method [26,27] | Lateral Scanline<br>Longitudinal Scanline |

Multi-Object Tracking (MOT) is vital for many applications in computer vision and has been extensively investigated. The object tracking methods can be classified into two types: online and offline tracking. The online tracking uses only current and previous frames. The long-term movements are embedded into a state-space for memorization [34-36]. The offline tracking is based on global optimization algorithms using a collection of time-series information of desired objects [37-39]. The recent deep learning approaches have gained tremendous momentum and successfully enhanced the performance for MOT, including Siamese Networks, Attention and Transformer, and Recurrent Neural Networks [40-43]. Some other practices consider the tracking issues from the mathematic formulation based on data association or machine learning models to extract trajectory features for clustering [44-46].

### *C. Scanline Method*

The Scanline method stems from the Spatial-Temporal Slice (STS) structure used in computer vision literature. Early spatial-temporal slice methods were introduced to solve the structure from motion (SFM) problem for cameras on moving platforms [47][48]. Later, the STS method was also used for object and pedestrian detection [49][50]. In transportation research, the technique is named as scanline method, which is a set of pixels that can capture object movements on the user-selected roadway from the video image. After stacking scanline pixels together over continuous frames, *Spatial-Temporal Map (STMap)* is obtained. On STMap, the horizontal axis shows the time progression, and the vertical axis contains distance information.

Two types of scanlines are used in traffic detection, lateral and longitudinal scanlines. The lateral scanline is a cross-section scanline across a lane, whereas the longitudinal scanline is along traffic direction. The lateral scanline method was intended primarily for traffic counting [51] and speed measurement [52]. The longitudinal scanline method was used for vehicle tracking [53][54] and detection [55]. However, most previous scanline methods were only used to estimate macroscopic parameters such as traffic count, headway, and spot speed.

A recent study [56] developed a High Angle Spatial-Temporal Diagram Analysis (HASDA) model to generate high-resolution (0.1s) vehicle trajectories using the longitudinal scanline. The HASDA model includes three major steps, the generation of STMap, the extraction of the pixel trajectories from STMap, and the coordinate transformation of the pixel to physical distance. The HASDA model mainly relies on image processing techniques such as background subtraction, shadow removal, edge detection, and morphological image operations. Some of the image processing techniques, such as edge detection, are vulnerable to noises. LiDAR data were combined with the scanline method to facilitate the calibration process in a later paper [57].

### D. Dynamic Mode Decomposition (DMD)

Dynamic Mode Decomposition is a data-driven analytic method that integrates Fourier transforms and singular value decomposition (SVD). DMD method was first introduced by Schmid [58] to extract meaningful spatial-temporal coherent structures that dominate dynamic activities in fluid mechanics. DMD method conducts the eigen decomposition of spatial-temporal coherent structures [59], therefore reducing the dimensions of complex systems efficiently without losing accuracy [60]. DMD methods have gained traction in many application areas such as fluid dynamics, video processing, control, epidemiology, and financial models. The DMD algorithm seeks to find the best fit between the following two matrices.

$$X = \begin{bmatrix} | & \cdots & | \\ x_1 & \ddots & x_{m-1} \\ | & \cdots & | \end{bmatrix} \quad , \quad X' = \begin{bmatrix} | & \cdots & | \\ x_2 & \ddots & x_m \\ | & \cdots & | \end{bmatrix} \qquad (1)$$

where $x_k$ $(k = 1, .., m)$ is a vector represents the dynamic system state at time interval $k$, $X$ matrix represents the prior states from intervals 1 to $(m-1)$, and $X'$ matrix represents the posterior states from intervals 2 to $m$.

Matrix $X$ and $X'$ are linked by a linear operator $A$:

$$X' = AX \qquad (2)$$

Our goal is to find the matrix $A$ that represents the evolution of system states. A well-studied least square estimation problem is formulated

$$\hat{A} = \underset{A}{\operatorname{argmin}} \|X' - AX\|_F^2 \qquad (3)$$

where $\hat{A}$ is an estimator of matrix operator $A$, which is computed by minimizing the Fresenius norm $\|X' - AX\|_F^2$.

$\hat{A}$ is governed by the least-square optimization

$$\hat{A} \approx X'X^{\dagger} \qquad (4)$$

Where $X^{\dagger}$ is obtained by using Moore-Penrose pseudoinverse.

For the DMD algorithm, instead of solving the matrix operator $A$ directly, $A$ is often solved by the eigendecomposition of $A$ after the proper orthogonal decomposition (POD).

***Step 1.*** Factorization of matrix X using SVD.

$$X \approx U\Sigma V^* \qquad (5)$$

Where $U$ and $V$ are column and row orthonormal bases of matrix X, $*$ denotes the complex conjugate transpose. Columns in $U$ are orthonormal, such that $U^*U = I$; Likewise, $V^*V = I$.

***Step 2.*** Reduce the dimension of A and obtain of $\tilde{A}$ by projecting A onto $U_r$ :

$$\tilde{A} = U_r^* A U_r = U_r^* X' V_r \Sigma_r^{-1} \qquad (6)$$

***Step 3.*** Compute eigenvalue of $\tilde{A}$:

$$\tilde{A}W = W\Lambda_r \qquad (7)$$

where the column of $W$ are eigenvectors and $\Lambda_r$ is the diagonal matrix of eigenvalues $\lambda_j$.

***Step 4.*** The eigenvector of A can be reconstructed using

$$\Phi = X'V\Sigma^{-1}W \qquad (8)$$

The above methods significantly reduce the complexities of the regression problem of estimating the full matrix A into calculating elements in the diagonal and sparse matrices. The DMD method can be considered a robust principal component analysis (PCA) with high computational efficiency. The eigenvalues of matrix $A$ can indicate the time evolution of dominant modes [61]. The way to stack state vectors together into a large matrix and identify its coherent structure is also named Method-of-Snapshots by Sirovich [62].

## III. Methodology

### A. STMap Generation

As illustrated in Figure 1, STMap is generated by stacking longitudinal scanlines $(l_1, l_2, l_3, \dots, l_m)$ frame by frame to form a three-dimensional matrix $S^{n \times m \times 3}$, where $n$ denotes the number of pixels per scanline, $m$ is the number of video frames, and 3 indicates the R-G-B channels.

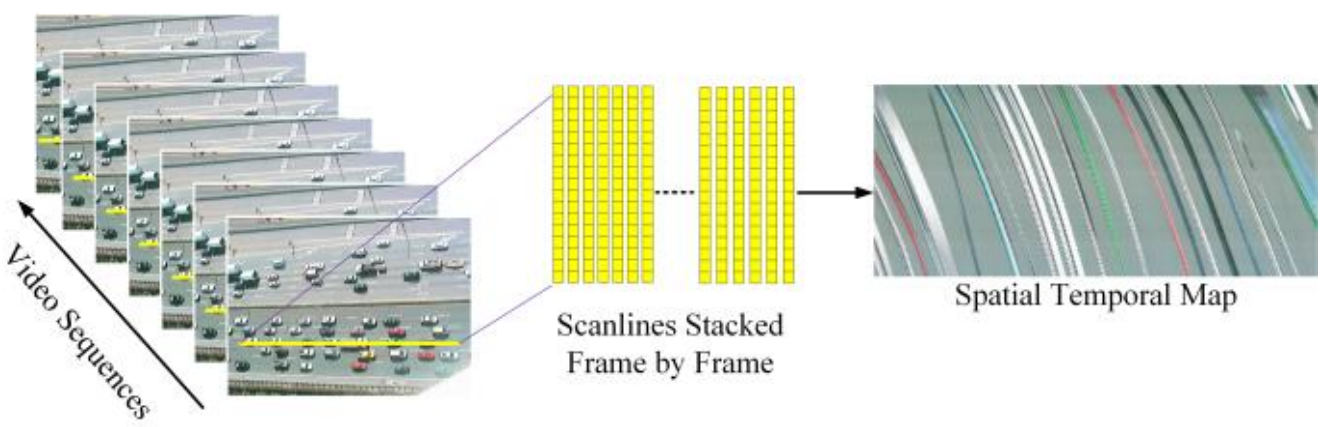

Figure 1 STMap Generation Using the Snapshot Vector

The color pixels moving simultaneously in the STMap represent a unique vehicle passing along the pre-defined scanline. Our purpose is to segment each vehicle strand from the STMap to detect trajectories.

### B. Dynamic Mode Decomposition (DMD) for STMap Segmentation

The scanline pixel values at each frame can be considered the state of traffic dynamic at a particular timestamp. Traffic state of $l_{x+1}$ at the time $(x+1)$ is assumed to relate to the previous traffic state of $l_x$ by linear time-dependent operator A, which reflects the time evolution of scanline pixels.

$$l_{x+1} = A\, l_x \qquad (9)$$

The STMap can be formulated as follows:

$$S = \begin{bmatrix} | & \cdots & | \\ l_1 & \ddots & l_{m-1} \\ | & \cdots & | \end{bmatrix} \quad , \quad S' = \begin{bmatrix} | & \cdots & | \\ l_2 & \ddots & l_m \\ | & \cdots & | \end{bmatrix} \qquad (10)$$

Where $S$ is the prior STMap, $S'$ is the posterior STMap. $S'$ has a one-frame shift from $S$. The relationship between $S$ and $S'$ becomes

$$S' = AS \qquad (11)$$

where the matrix $A$ describes the time-differencing operation. The DMD mode that contains spatial information is an eigenvector of $A$. And each DMD mode corresponds to an eigenvalue of $A$. By finding the eigenvectors and eigenvalues of the matrix $A$, we obtain the DMD mode $\Phi$.

$$A\Phi = \Phi\Lambda \qquad (12)$$

The column of $\Phi$ are eigenvectors comprising of the dominant mode $\emptyset_j$ and $\Lambda$ is the diagonal matrix of eigenvalues

$\lambda_j$. The STMap can be reconstructed using first $k^{th}$ modes, where $k \leq \min(n, m)$.

$$STMap \approx \Phi \mathrm{B} \mathcal{V} = \left(\begin{bmatrix} \emptyset_{11} & \cdots & \emptyset_{1k} \\ \vdots & \ddots & \vdots \\ \emptyset_{n1} & \cdots & \emptyset_{nk} \end{bmatrix}\right)\left(\begin{bmatrix} b_1 & \cdots & 0 \\ \vdots & \ddots & \vdots \\ 0 & \cdots & b_k \end{bmatrix}\right)\left(\begin{bmatrix} 1 \ \lambda_1 & \cdots & \lambda_1^{m-1} \\ 1 \ \vdots & \ddots & \vdots \\ 1 \ \lambda_k & \cdots & \lambda_k^{m-1} \end{bmatrix}\right) \quad (13)$$

where $\Phi$ contains the dominant modes from the STMap, matrix B is the matrix of amplitudes. $\mathcal{V}$ is the Vandermonde matrix representing the time evolution of DMD modes. This function is illustrated in Figure 2.

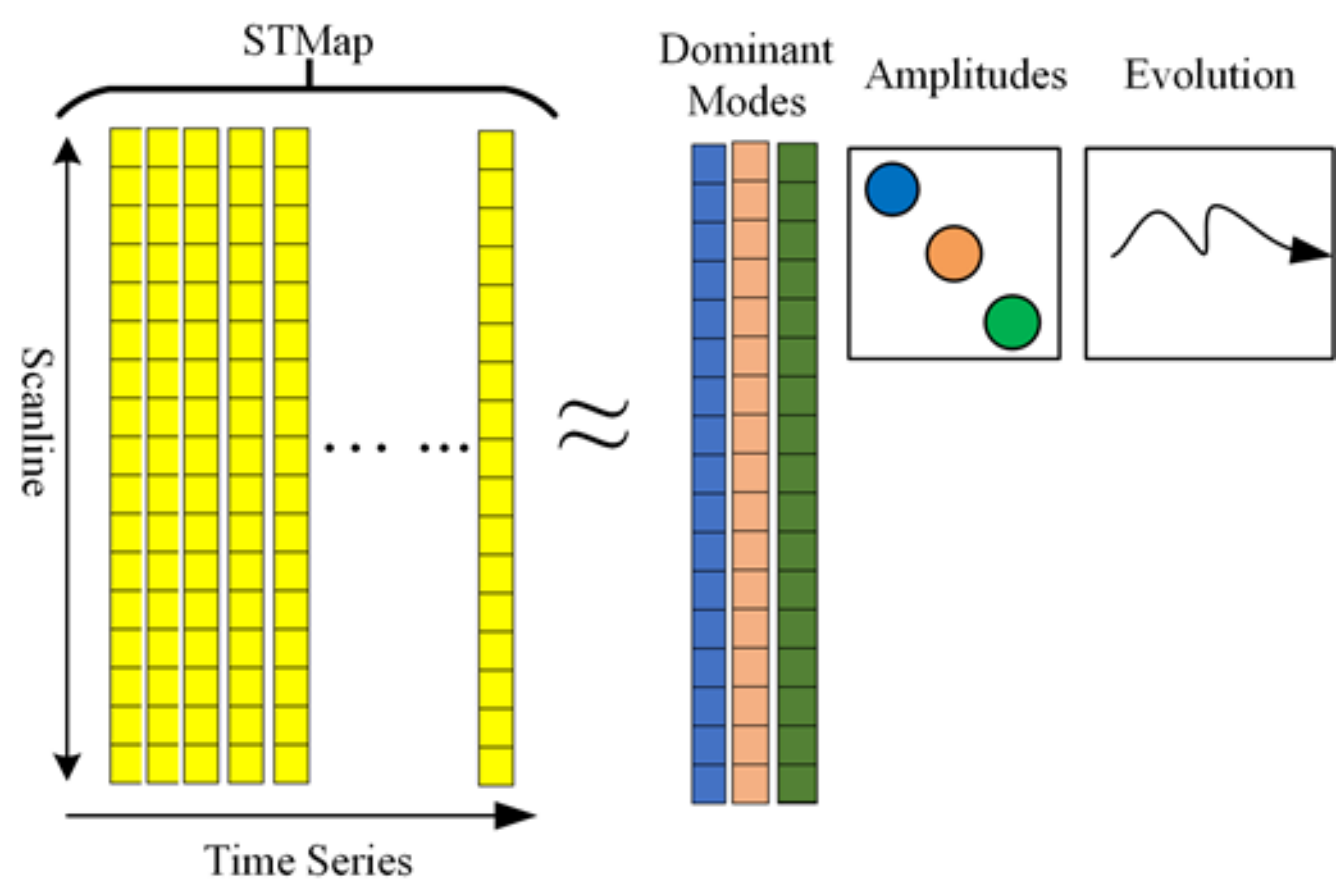

Figure 2 Decompose Spatial-Temporal Map into Dominant Modes

A scanline vector $l_t$ at frame $t \in 1, \ldots, m$ can be estimated as follows:

$$\widetilde{l_t} = \sum_{j=1}^{k} b_j \emptyset_j \lambda_j^{t-1} \quad (12)$$

where $b_j$ is amplitude, $\emptyset_j$ is each DMD mode, and $\lambda_j^{t-1}$ is the time evolution of each mode.

Let $t = 1$, which represents the initial state of scanline as follows.

$$\widetilde{l_1} = \sum_{j=1}^{k} b_j \emptyset_j \quad (13)$$

The matrix B can then be estimated as a least-square problem using the first scanline $l_1$ as an initial state.

$$\tilde{B} = \underset{B}{argmin} \, \|l_1 - \Phi B\| \quad (14)$$

Any DMD mode that does not change in time will have an eigenvalue $\lambda_j = 1$, which forms the background of the STMap.

In the STMap, the background pixels are highly correlated between neighboring columns, suggesting the low-rank structure within the STMap. Therefore, the DMD algorithm separates background and foreground by decomposing the STMap into low-rank (background) and sparse (foreground) components.

$$S_{DMD} = background + forground = \sum_p b_p \phi_p \lambda_p^{t-1} + \sum_{j \neq p} b_j \phi_j \lambda_j^{t-1} \quad (15)$$

where $|\lambda_p| = 1$. $t \in 1, \ldots, m$ indicates the frame number.

As displayed in Figure 3, the background is time-independent and has a minimal DMD eigenvalue ($\approx 10^{-6}$) compared with the foreground modes' eigenvalues. Figure 3 (D) plots the oscillations of selected methods over time. The amplitude of the background mode with the lowest frequency indicates the least variations over time.

As shown in Figures 3 (A) and 3 (B), DMD's foreground detection results are not perfect for trajectory extraction on STMap. However, they are clear enough to preprocess and generate training data for deep learning models.

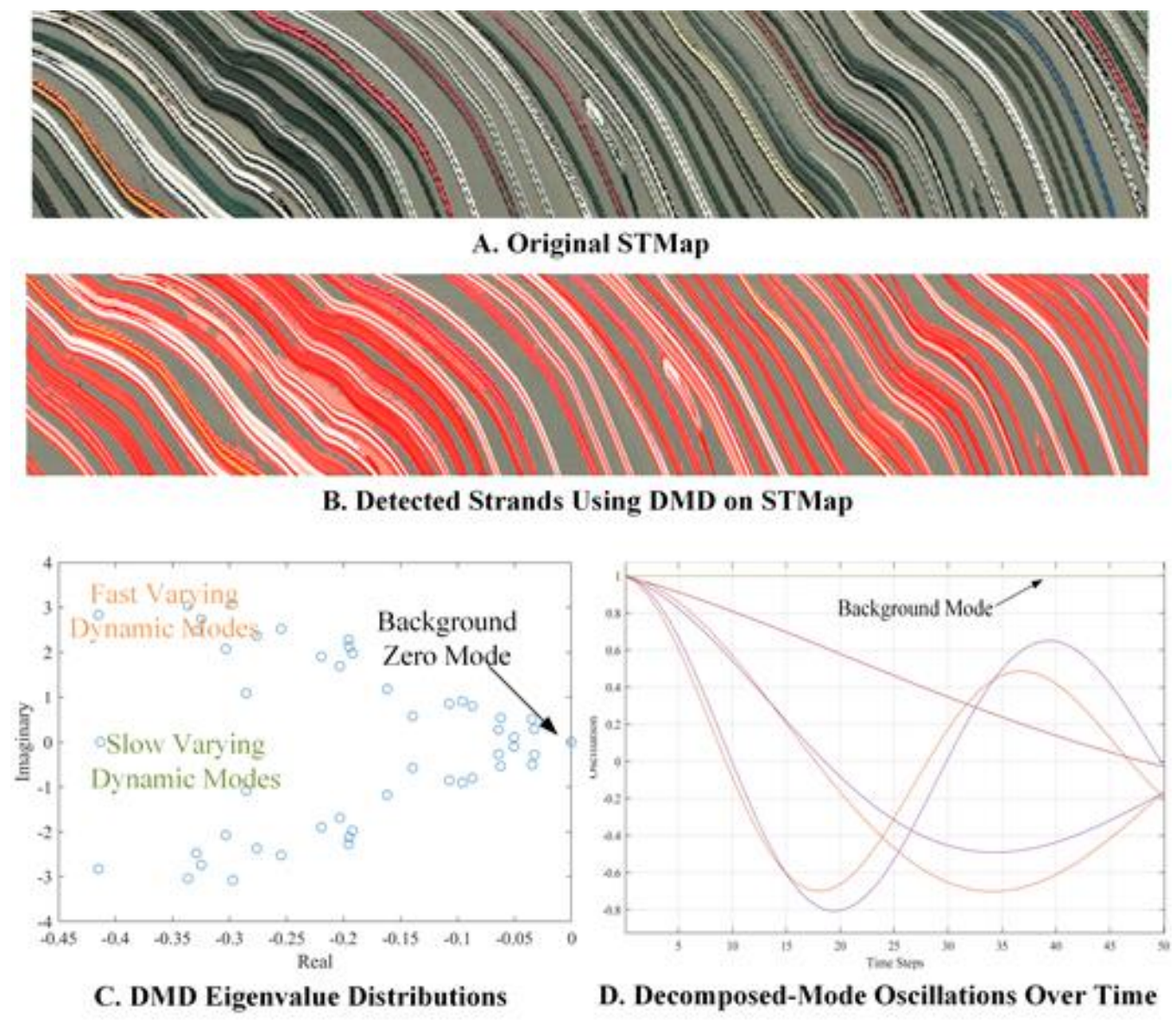

Figure 3 STMap and Its Dynamic Modes

### C. *Res-UNet+ Model for STMap Segmentation*

The Res-UNet+ model uses the ResNet block as the backbone and further increases its performance by modifying the decoding layers. In the encoding process, the ResNet blocks replace the original encoders in the UNet model. The interconnections between encoding and decoding layers were designed to reduce the semantic gap. We added the intra-connections among different levels of decoding stages. Many segmentation studies [63-65] show that features learned from different scales often entail different information. Lower-level layers capture the boundary of objects, whereas higher-level layers explore the localization of the targeted objects. In the vanilla UNet architecture, there are only interconnections between the corresponding level of encoders and decoders. To enable lower-level decoder information to pass to higher-level decoders, we concatenate all the decoder layers to localize better and learn representations in the segmentation network. The multi-scale intra-connection and depth of each layer are shown in Figure 4, and the network is named Res-UNet+.

The encoder layer using two branch ResNet backbone, including upper branch and lower branch, is defined as the following:

$$H_l^i = ReLU(id(H^{i-1}) + F_1(H^{i-1})) \quad (16)$$
$$H_u^i = F_2(H_l^i) + F_1(H_l^i) \quad (17)$$
$$H^i = ReLU(H_u^i) \quad (18)$$

where $H_l^i$ is lower branch output of $i^{th}$ layer encoder. $H_u^i$ is upper branch output of $i^{th}$ layer encoder. $H^i$ is the $i^{th}$ layer output. $F_1 = BN(\mathrm{Conv3}(\mathrm{ReLU}(\mathrm{BN}(\mathrm{Conv3}(\mathrm{x})))))$, and Conv3 is 3 by 3 convolution operator. $F_2 = BN(\text{Conv3(x)})$, BN is Batch Normalization.

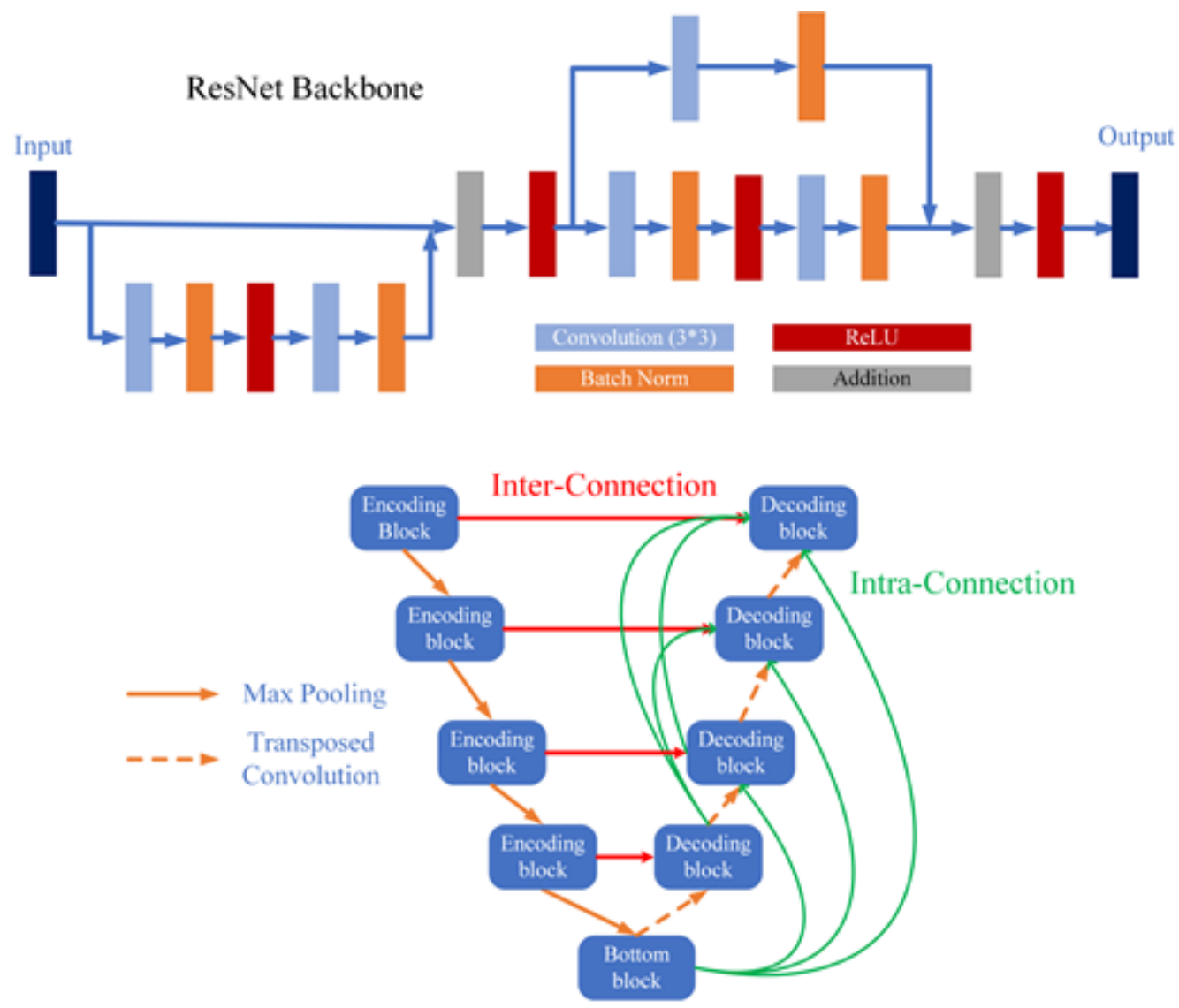

Figure 4 Proposed Res-UNet+ Topological Structure and Building Block

The decoder layer with intra-connections between different levels is described as the following:

$$L^i_{de} = \begin{cases} A^2([L^n_{en}, A_T(L_{bottom})]), & i = n \\ A^2\left(\left[A(L^i_{en}), A_T\left(L^j_{de}\right)^{j=n}_{j=i-1}\right]\right), & i < n \end{cases} \quad (19)$$

where $L^i_{de}$ is the $i^{th}$ decoder layer output, $L^i_{en}$ is the $i^{th}$ encoder layer output, $[\cdot]$ represents the depth concatenation operation, $A(\cdot)$, $A_T(\cdot)$, and $A^2(\cdot)$ represents respectively convolution, transposed convolution, and twice operation of convolution, followed by the ReLU activation, $L_{bottom}$ is the bottom bridge layer output.

## IV. Model Implementation

### 1) Baseline Models in the Study

Mainstream image semantic segmentation models as baseline models are considered as follows:

***ResNet-18/ ResNet-50:*** The building block for ResNet includes the main branch that contains convolution, batch normalization, and ReLu layer consecutively, as well as a residual connection that bypasses the main stem to allow the gradients to flow more easily. In this paper, we tested the 18-layer and the 50-layer ResNet architecture as reference models. The left branch of Figure 4 illustrates a similar ResNet structure used in the proposed model.

***UNet:*** The vanilla UNet model with encoding and decoding stages is also used as a reference model. The skip connection is comprised of two sets of convolutions and ReLU layers. The vanilla UNet has a U-Shaped structure similar to Figure 4 but with a more straightforward left branch and without the intra-connected decoding layers defined in the proposed model.

***Res-UNet:*** For the Res-UNet model, we did not add intra-connection to integrate the information from all encoder layers. Our Res-UNet architecture reformed the UNet model architecture by replacing the original encoder layer with a two-branch ResNet block.

***Fully Convolutional Network (FCN):*** The FCN model is an end-to-end encoder-decoder semantic segmentation neural network. The encoder-decoder architecture was inherited by almost all subsequent segmentation models [66].

***DeepLabv3+:*** DeepLab model was also built on the encoding-decoding framework, adopting the Xception model and the depthwise separable convolution achieve a faster and stronger encoder-decoder network [67].

***SegNet:*** Another pixel-wise segmentation neural network is SegNet, which uses 13 convolutional layers topologically similar to VGG16 as the encoder. Their decoder layer uses pooling indices computed in the corresponding encoder layer to perform non-linear upsampling [68].

### 2) Dataset and Augmentation

Unlike other data labeling process, which is highly specialized and requires a lot of expertise and experience, a single person will suffice to complete the task of labeling hundreds of STMap using the mentioned DMD methods. This is one of the advantages of using the STMap method, as there is no need to collect vehicle images from all possible scales, parts, angles, colors, or shapes. Furthermore, the patterns of vehicle strands compared to background pixels are easy to be partitioned.

In this study, the STMap training dataset was created using four 15-minute NGSIM I-80 videos. We obtained 20 STMaps from 20 lanes, which were then cropped into 1000 512*512 images. Finally, we augmented the 1000 images to get more datasets by rescaling, shearing, and translating the generated 512*512 images. Since the vehicle strands in the STMap extend from the top left to the bottom right, we do not need to use the rotation transformation in the data augmentation process.

### 3) Implementation Details

We programmed the proposed algorithms with image processing and Deep Learning Toolbox in MATLAB. The pre-trained model parameters were retained as the initialization values. The stochastic gradient descent with momentum is used to optimize the proposed network, and the batch size is set to 3 images each time due to the GPU memory constraints, and the learning rate is initialized as 0.05. All the models were trained on GeForce GTX 1060 (6 GB memory). The maximum epoch is 10, equivalent to a training time of around 5 hours. Once trained, the proposed model takes 2 minutes to process the STMap from a 15-minute video.

All training sets were shuffled and partitioned into 60% training, 20% testing, and 20% validation. Two classes are defined in the pixel classification layer, vehicle strands, and background. Our goal is to correctly classify all pixels belonging to the vehicle strands from STMap and then extract the vehicle strands boundary.

### 4) Trajectory Extraction

After the segmentation and obtaining the binary masks for all vehicle strands from STMap, the following step is to extract the vehicle pixel trajectory. The vehicle trajectory is acquired using the lower boundary of each vehicle strand. In MATLAB, it's just a one-line function bwperim(), making the method super-efficient.

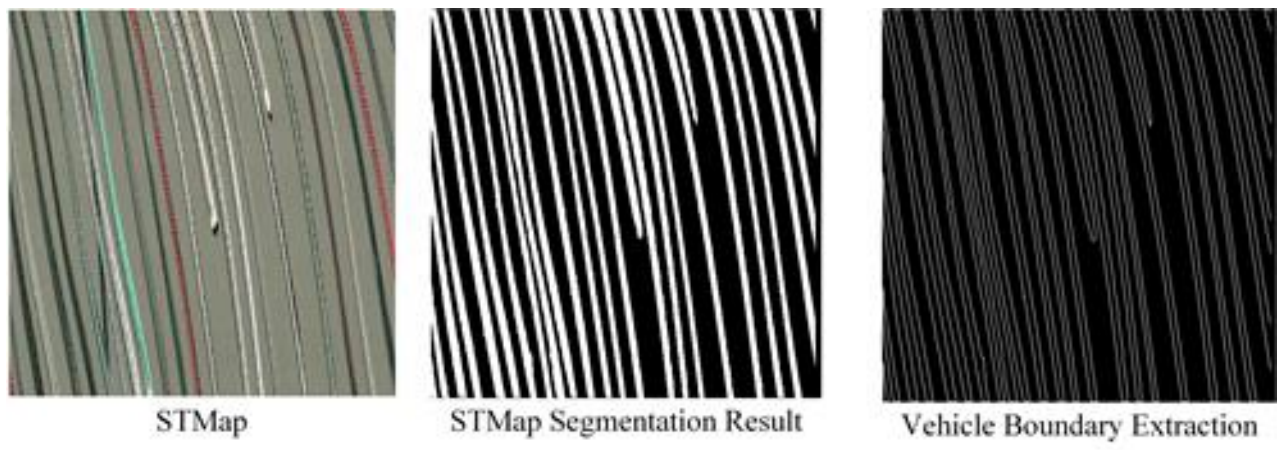

Figure 5 Extracting Vehicle Trajectory from STMap Segmentation

The pixel trajectories from STMap can then be converted into pixel movements on the original video. Then the video trajectory will be converted into NGSIM Local-y coordinates using the method in [56].

### 5) *Performance Metrics*

Three main performance metrics were used to quantitatively assess the segmentation model performance, accuracy, Jaccard coefficient, and BF Score (Boundary F1 Score).

Accuracy (Acc) is calculated as the percentage of correctly classified pixels to the total number of pixels for each class regarding the ground truth data.

$$Acc = \frac{TP}{(TP + FN)} \quad (20)$$

where TP is the count of true positives; FN is the count of false negatives. However, Acc is a primary metric and needs to use in junction with other metrics for a complete evaluation.

The Jaccard coefficient is used to measure the similarity between two sets A and B (also known as intersection over union or IoU) is defined as:

$$J(A, B) = \frac{|A \cap B|}{|A \cup B|} \quad (21)$$

Jaccard coefficient is an overlap index that quantifies the agreement between two segmented image areas. The Jaccard coefficient can also be expressed in terms of true positives (TP), false positives (FP), and false negatives (FN) as:

$$J(A, B) = \frac{TP}{(TP + FP + FN)} \quad (22)$$

The third performance metric is BF Score, which computes the contour-matching score between model-predicted results and ground truth data. As shown in the following equation, the BF Score is defined as the harmonic mean between precision and recall values to decide whether the points on the boundary have been matched.

$$BF = 2 * \frac{precision * recall}{(recall + precision)} \quad (23)$$

The error measure for the trajectory detection results is the mean absolute error (MAE) of all trajectory points.

$$MAE_o = \frac{1}{N}\sum_{t=1}^{N}|\hat{y}_o(t) - y_o(t)| \quad (24)$$

Where $o$ is the trajectory index, $y_o(t)$ and $\hat{y}_o(t)$ are the actual and model-estimated local_y location at time t, respectively, $MAE_o$ is the Mean Absolute Error between the ground truth trajectory and the estimated trajectory by averaging all distance discrepancies within the common time window. If the mean absolute error is below a pre-determined threshold (15 ft in this study), we will consider the detected trajectory as a true-positive. Otherwise, it will be considered a false-positive result.

## V. Experimental Design

The video data used in this study was from the NGSIM I80-1 dataset, which was recorded from 4:00 p.m. to 4:15 p.m. on April 13, 2005, on I-80 in Emeryville, California. The direction of traffic flow recorded was northbound. Each camera watched vehicles passing through the study area from the roof of a 30-story building adjacent to the freeway. Five lanes from four cameras were used in the study, including a high-occupancy vehicle (HOV) lane, as shown in Figure 6.

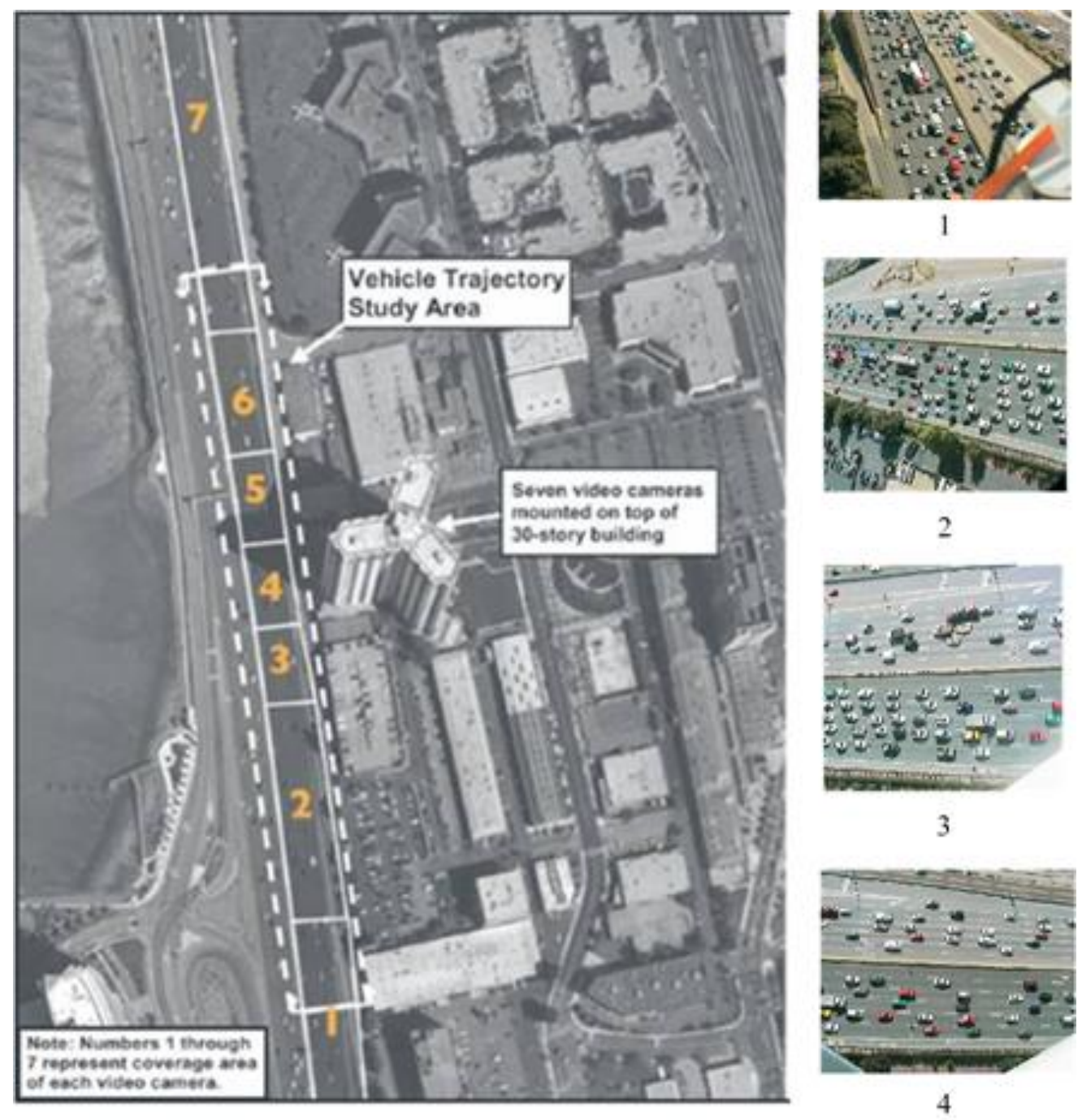

Figure 6 NGSIM I-80 Survey Area and 4 Study Cameras [1]

## VI. Model Evaluation

In this section, we show the STMap trajectory detection results using proposed deep learning models. We will discuss both advantages and disadvantages of the proposed models for the STMap trajectory segmentation quantitatively and qualitatively. We will also compare the trajectory detected by the proposed model against the NGSIM trajectory.

### A. *STMap Segmentation Evaluation*

Figure 7 compares the proposed Res-UNet model and Res-UNet+ model with the baseline models using selected performance metrics on the testing dataset. Global accuracy is the number of correctly classified pixels over the total number of pixels. The mean accuracy is the average accuracy for each class. Mean IoU is the average IoU score across all classes. Finally, the weighted IoU is the IoU score weighted by the number of pixels for each class in the image. Using weighted metrics is to reduce the impact of imbalanced classes.

As shown in Figure 7, the UNet model family outperformed the ResNet model family in the segmentation task. The vanilla UNet produced better results with only 70 layers than ResNet-50 with 206 layers because of the interconnection between the encoding and decoding stage. A fully convolutional Neural Network (FCN) also yielded desirable segmentation results. Deeper ResNet-50 outperforms the less deep ResNet-18 model within the same model family. Intra-connections between decoding layers are efficient, as Res-UNet+ has better performance than Res-UNet or UNet built with fewer connections. Res-UNet+ outperforms all the rest networks.

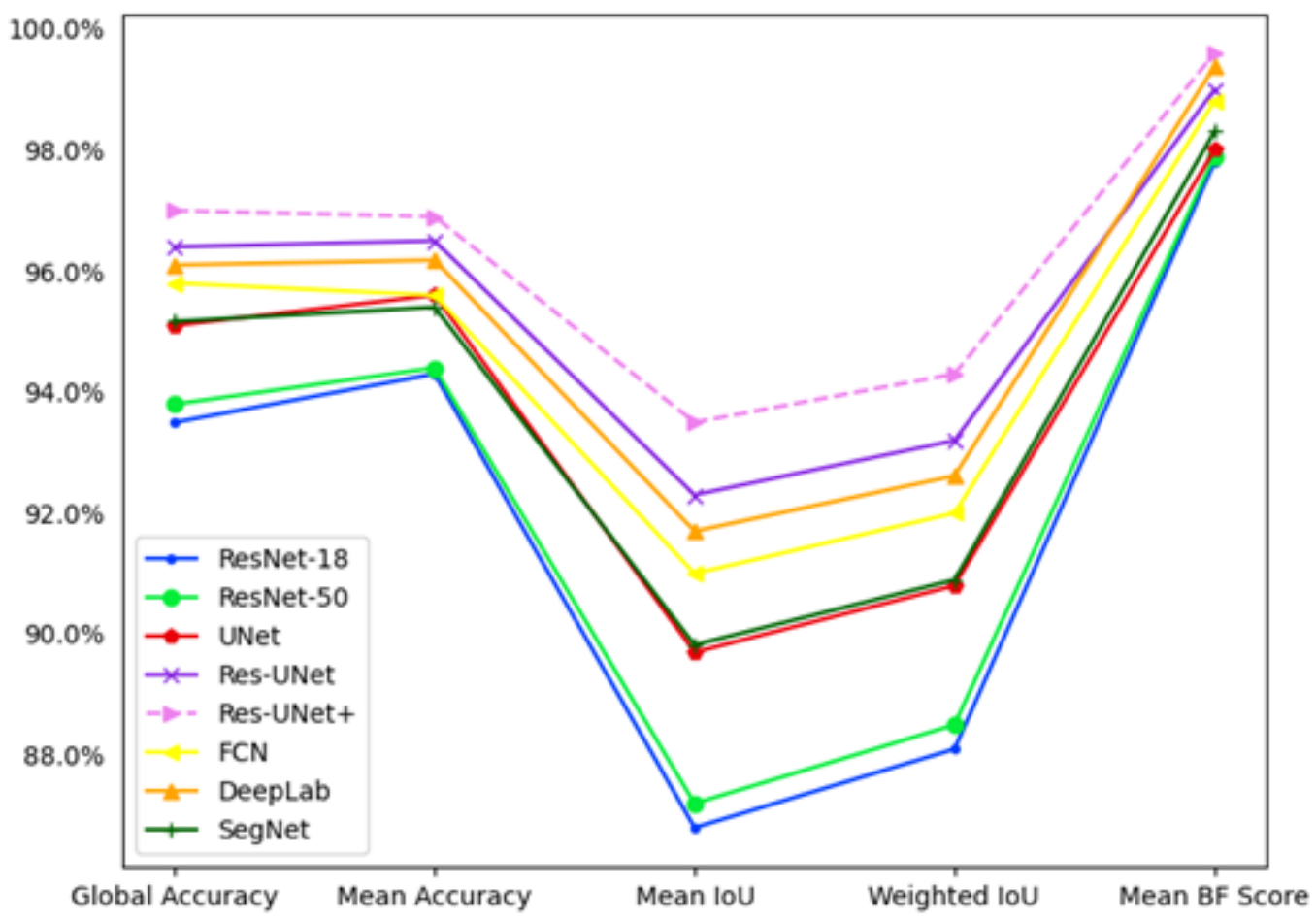

Figure 7 Model Performance Comparison with Baseline Models

Figure 8 performs a detailed comparison of the trajectory-level performance between the proposed model and the reference models by inspecting the vehicle segmentation results (in yellow stripes) from a representative STMap section impacted by shadows and occlusions. The STMap shows four major shadow stripes created by vehicles from neighboring lanes. All tested models can distinguish the vehicle strands from static noises caused by lane markers. The proposed Res-UNet+ model produces the cleanest boundaries for detected vehicle strands. Res-UNet+ is also more robust against vehicle shadows than other deep learning models tested (black cycles). Most of those models classify shadows as vehicle strands in difficult shadow situations, which will cause false detections. The false detected shadows stitch multiple vehicle strands together will result in missed detections. The proposed Res-UNet+ model was able to extract strands with fewer shadow fragments. Some small shadow pixels are detected, but those can be easily filtered out since they are isolated with a small number of pixels.

In the oscillation scenario, the ResNet model family creates holes within vehicle strands or overlaps with nearby vehicle stands (Red cycles). The Res-UNet+ model generates the best overall segmentation results.

### B. *Trajectory Level Evaluation*

This section further evaluates the trajectory-level performance. Three detecting cases are considered, including true positive (TP), false positive (FP), and false-negative (FN). True positive rate (TPR) and false-positive rate (FPR) as used to evaluate Type I and II errors.

$$\mathrm{TPR} = \mathrm{TPs}/(\mathrm{TPs} + \mathrm{FNs}) \quad (26)$$

$$\mathrm{FPR} = \mathrm{FPs}/(\mathrm{TPs} + \mathrm{FPs}) \quad (27)$$

The proposed model is compared with the prior HASDA model developed by the research team [56] in Table II.

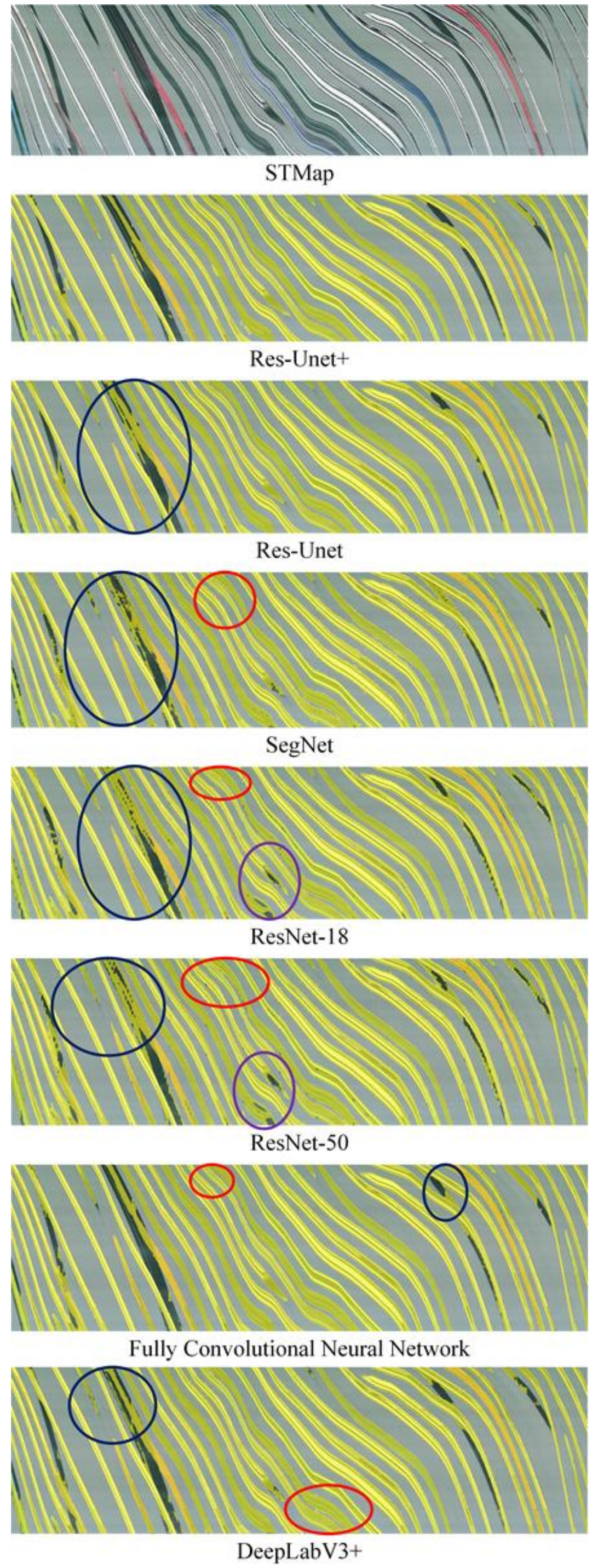

Figure 8 Segmentation Results Using Proposed and Baseline Models (Black Cycle: Shadows; Red Cycle: Overlapping; Purple Cycle: Holes)

TABLE II
MODEL PERFORMANCE AT TRAJECTORY LEVEL

| | Lane 1 | Lane 2 | Lane 3 | Lane 4 | Lane 5 |
|---|---|---|---|---|---|
| **Res-UNet+ Model** | | | | | |
| Positive Detect Rate | 97.1% | 98.9% | 98.1% | 98.0% | 96.5% |
| False Detect Rate | 3.2% | 2.7% | 3.2% | 4.2% | 3.0% |
| **HASDA Model** | | | | | |
| Positive Detect Rate | 96.5% | 96.9% | 94.1% | 89.58% | 92.3% |
| False Detect Rate | 5.8% | 6.9% | 2.4% | 9.5% | 3.5% |

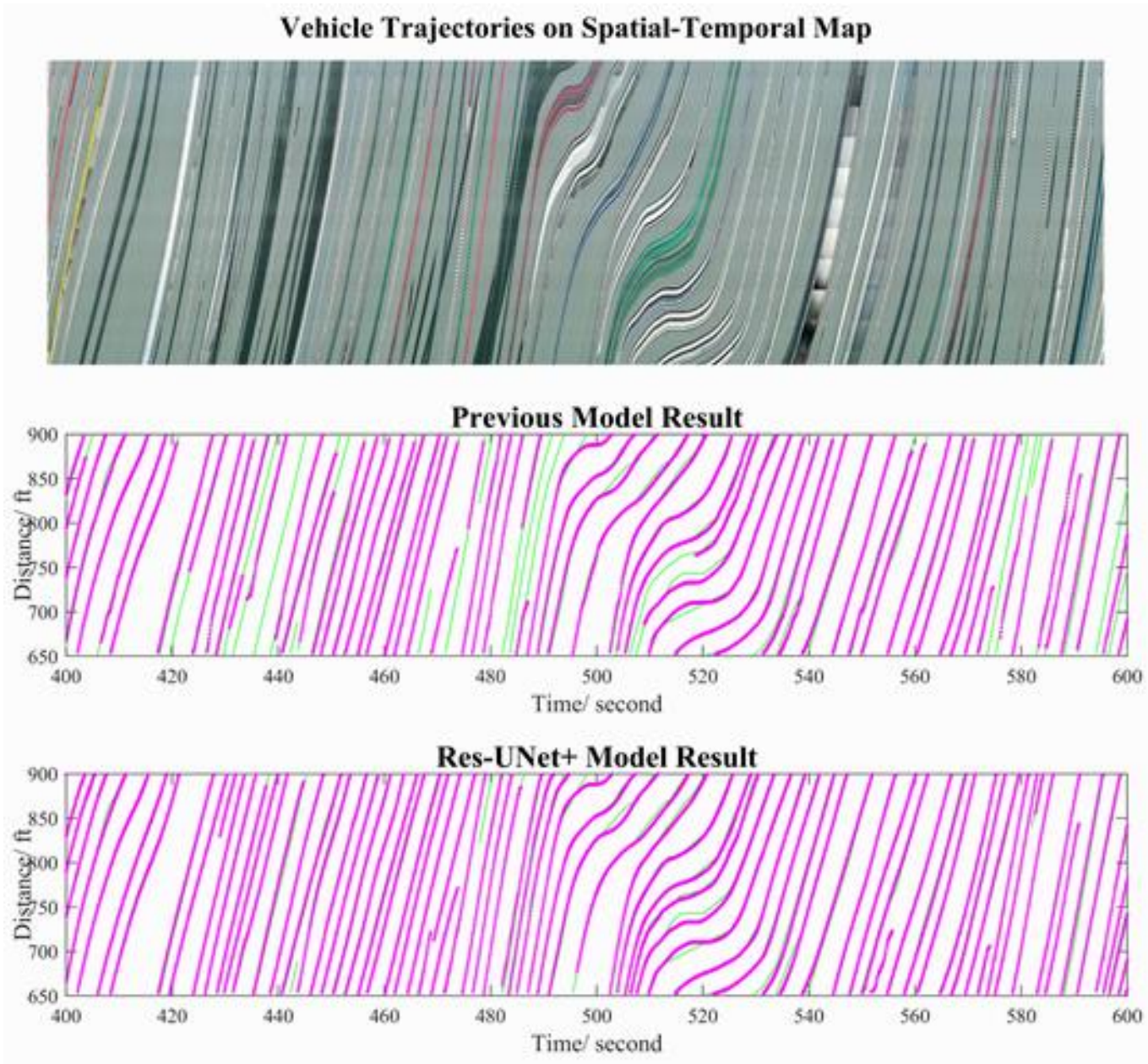

Figure 9 HASDA Model [56] and Res-UNet+ Model (Purple: Reconstructed Trajectories; Blue: NGSIM Trajectory)

Figure 9 shows that the deep learning model is more robust than the previous HASDA model under the influence of shadows. NGSIM trajectories are used as the reference to provide a microscopic inspection of the trajectory detection results. The deep learning model has greater accuracy and completeness with its capability to capture different semantic characteristics levels. For the stop-and-go traffic, the deep-learning model detection results captured the oscillations very faithfully.

## VII. NGSIM Data Reconstruction

Despite its importance and frequent use in literature, NGSIM data have critical quality issues. Therefore, data cleaning becomes a prerequisite before using NGSIM data to study traffic flow characteristics or models. The previous filtering techniques relying on statistical filtering or smoothing can only improve trajectory data marginally. The proposed models were used to identify NGSIM data quality issues and then reconstructed the dataset completely. By processing the NGSIM videos from four cameras (1000 ft area), we thoroughly cleaned the datasets and achieved significant quality improvement.

As shown in Figure 10, the vehicle detection results and the NGSIM detection results are plotted on the raw NGSIM I-80 video. The blue lines are plotted based on the local-y and vehicle length data from NGSIM I-80 trajectory data. The red bars are the detected vehicle fronts by the proposed model. As shown in Figure 10, the NGSIM data have significant drifting issues. At the same time, the proposed model detects vehicle fronts very close to their actual positions in the video. The drifting problem of NGSIM data occurs more frequently when vehicles are joining or leaving a congested platoon. Figure 10 also shows a different NGSIM tracking error that vehicle 2144 was misidentified as vehicle 2143 after vehicle 2144 changed from lane 5 to lane 4. This type of error cannot be corrected with smoothing or filtering. The fixed vehicle trajectories for 2143 and 2144 in our reconstructed NGSIM data are shown in the lower subplot of Figure 10.

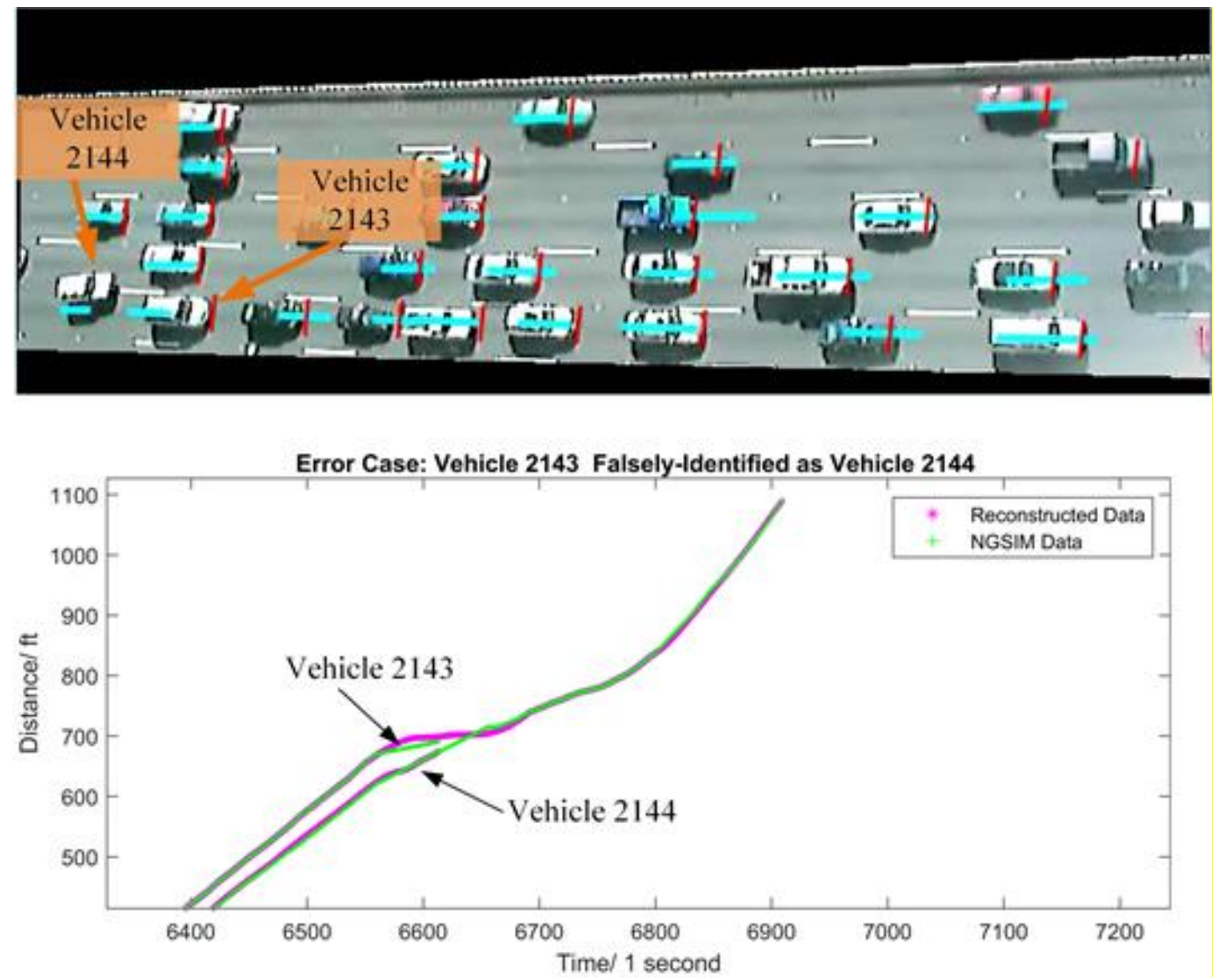

Figure 10 Video Validation of Reconstructed NGSIM data (Blue: NGSIM; Red: Detection results of the proposed model)

The second type of error in the NGSIM data is caused by the homography projection of raw video that assumes all objects are on the same ground plane The simplified 2D plane homography assumption during the video transformation can lead to significant self-occlusion issues. Figure 11 shows image projection error in NGSIM datasets for large vehicles. Due to the false assumption of the 2D homography projection that all objects are on the same ground, the NGSIM data often capture the off-ground features of large vehicles (e.g., trucks and buses). The self-occlusion can lead to significant positioning errors for large vehicles, as illustrated by the white bus from the HOV lane in Figure 11. However, the proposed model is applied directly to the raw NGSIM video data and only extracts features on the ground and can track the large vehicle accurately.

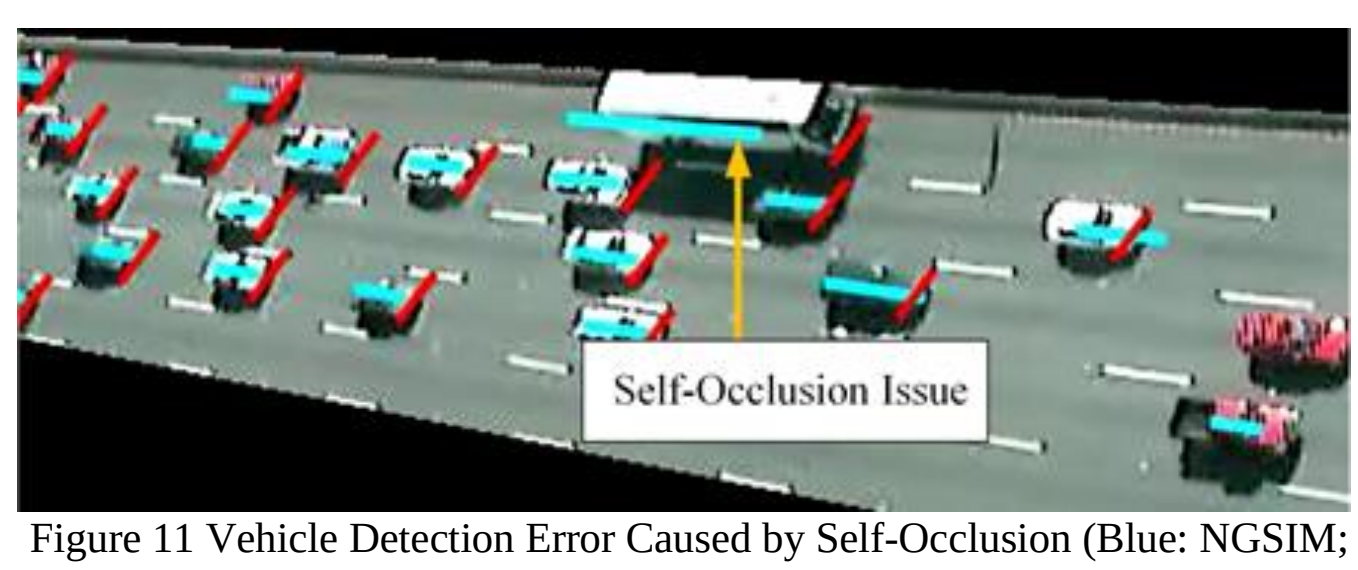

Figure 11 Vehicle Detection Error Caused by Self-Occlusion (Blue: NGSIM; Red: Detection results of the proposed model)

Figure 12 exemplifies typical errors in shockwaves in the NGSIM dataset caused by the unstable vehicle detection and tracking. We reversely plot the NGSIM trajectory data onto the STMap and found that the NGSIM data have issues at almost every shockwave condition. This may explain some of the calibration issues of microscopic car-following models based on NGSIM data [69]. The proposed model was able to generate trajectories much more consistent with the vehicle strands in STMap. Results from the proposed model and the raw NGSIM datasets are displayed side-by-side in Figure 12.

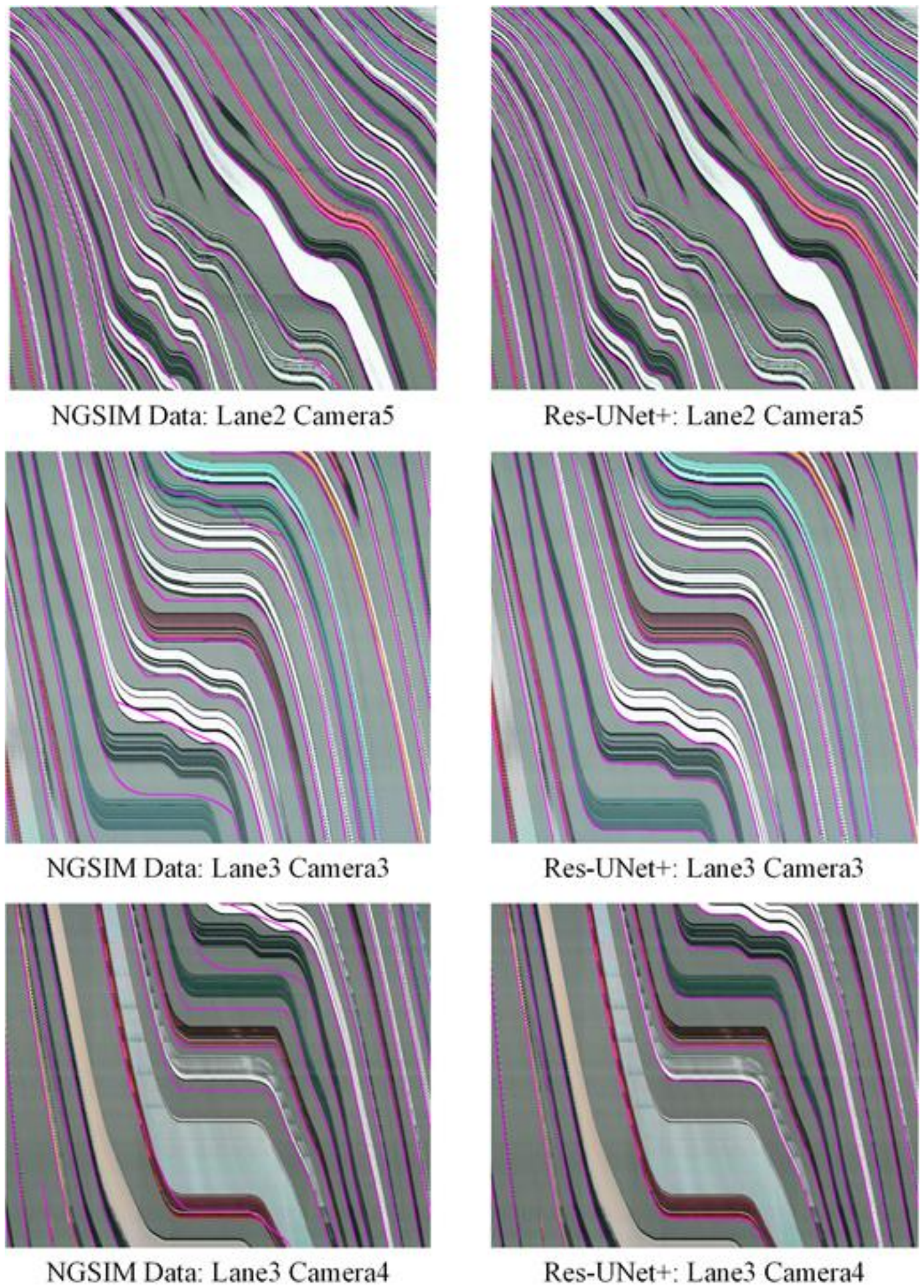

Figure 12 NGSIM Trajectories with Issues v.s. ResUNet+ Trajectories (Purple: Pixel Trajectory)

The original NGSIM video processing method is a multi-stage process that has led to some data quality issues. The proposed method can simplify the trajectory generation and validation process and improve data quality. All the validation videos can be found in [70]. The reconstructed NGSIM dataset is made openly available to the public. The code for the Res-UNet+ model and DMD model will also be publicly available.

## VIII. Conclusion

The proposed Res-UNet+ integrates two prevailing CNN structures to extract vehicle trajectories from the Spatial-Temporal Map (STMap). The STMap-based video analytics only analyzes the snapshot of a few scanlines from each video frame rather than the entire image, making it more computationally efficient than tracking-by-detection models that process the whole video frame by frame. The machine-learning-enhanced scanline methods addressed the issues of earlier image processing models (e.g., the HASDA model), especially against static noises, moving shadows, and vehicle occlusions. The data labeling process is semi-automated by using the DMD method.

The proposed Res-UNet+ model is evaluated with the NGSIM I-80 dataset and performs better than several baseline deep learning models. Running the proposed model is also computationally efficient, which can be deployed in parallelism for each scanline. The reliable trajectory results using STMap can fundamentally address the data quality issues caused by the limitations of the conventional tracking-by-detection paradigm. In response to the growing appeal for accurate trajectory datasets, this research reconstructed the NGSIM dataset using raw video data and solved nearly all error cases.

Future work of the research includes more extensive testing and sensitivity analysis under different traffic conditions (e.g., arterials, intersection, highway merging) and camera conditions (e.g., roadside CCTV cameras and Drones). In addition, the online version of the proposed algorithm will also be developed to support real-time intelligent transportation applications.

## IX. Appendix: notation

$X$, $X'$: Prior Snapshot Matrix, Posterior Snapshot Matrix.

$S$, $S'$: Prior STMap, Posterior STMap.

$A$, $\hat{A}$ and $\tilde{A}$: Linear Transformation Matrix, Least Square Estimator, and Reduced-rank Linear Matrix.

$x_{\mathrm{k}}$: One snapshot at column $k$.

$l_k$: Scanline of STMap at column $k$.

$\Phi$, $\phi_j$: Matrix of DMD modes, and single DMD mode $j$.

$\Lambda$, $\lambda_j$, Matrix of eigenvalues, and Single Eigenvalue $j$.

$B$, $b$: Diagonal Matrix of Amplitudes, Single amplitude.

$\mathcal{V}$, Vandermonde matrix.

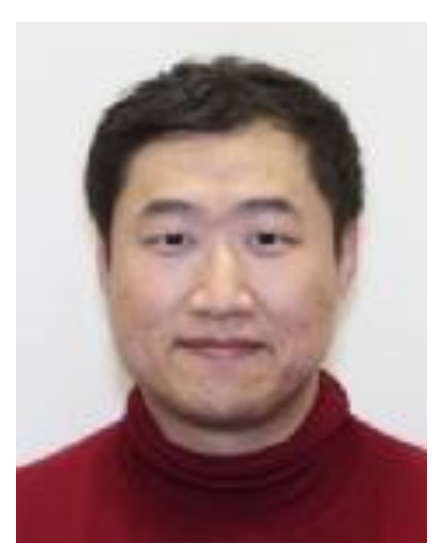

**Tianya Zhang** received BS degree from Beijing Institute of Technology, China, and earned master's degree from Texas A&M University, College Station, Texas, US. Currently, he is a Ph.D. candidate in the Department of Civil and Environmental Engineering from Rutgers University.

He has been involved in several ITS projects related to Automated Traffic Signal Performance Measures (ATSPMs), Connected and Automated Vehicle (CAVs) and Smart Mobility Testing Ground. His research using computer vision and LiDAR for vehicle trajectory detection has been published in Transportation Research Part C and Transportation Research Record.

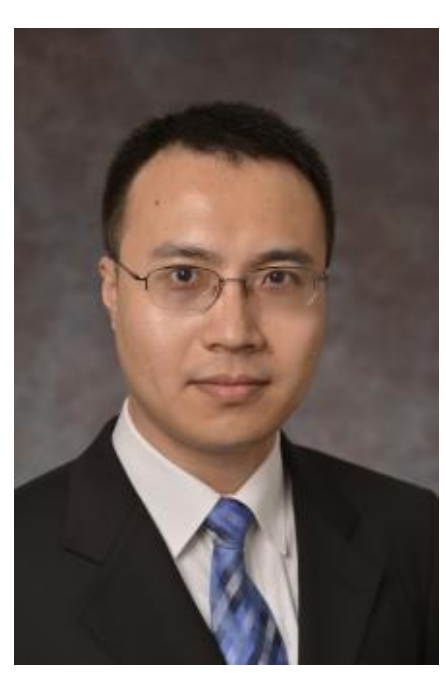

**Peter J. Jin** was an Associate Professor at Department of Civil and Environmental Engineering (CEE) at Rutgers, The State University of New Jersey. He received the BS degree in Automation at Tsinghua University, China. He received his MS and Ph.D. degrees in civil engineering from University of Wisconsin-Madison in 2007 and 2009 respectively. He worked at Center for Transportation Research, at the University of Texas at Austin as a postdoctoral fellow and research associate. He has more than 45 peer-reviewed journal publications and more than 70 conference papers. His research interests include transportation big data analytics, intelligent transportation systems, connected and automated vehicles (CAVs), and unmanned aerial vehicles (UAVs). He holds two patents in both UAVs and CAVs.